\documentclass[runningheads]{llncs}

\usepackage{eccv}

\usepackage{eccvabbrv}

\usepackage{graphicx}
\usepackage{booktabs}
\usepackage{bm}
\usepackage{multirow}
\usepackage{makecell}
\usepackage[table]{xcolor}
\usepackage[accsupp]{axessibility}  % Improves PDF readability for those with disabilities.

\usepackage{xcolor}
\usepackage[colorlinks=true, linkcolor=myblue, urlcolor=myblue, citecolor=myblue]{hyperref}
\definecolor{myblue}{RGB}{100, 149, 237}  % cornflower blue, soft and readable

\usepackage{orcidlink}

\begin{document}

\title{RAF: Reliability-Aware Fusion of Camera, LiDAR, and 4D RADAR for Robust 3D Object Detection in Adverse Weather} 

\titlerunning{RAF}

% Include the authors' ORCID for the camera-ready version, if at all possible.
\author{Heejun Park\inst{1}\orcidlink{0009-0001-1166-7623} \and
Jaeseok Jeong\inst{1}\orcidlink{0000-0002-9836-2979} \and
Kuk-Jin Yoon\inst{1}\orcidlink{0000-0002-1634-2756}}

\authorrunning{H. Park et al.}

\institute{Visual Intelligence Lab., KAIST \\
\email{\{parkhee.ticket, jason.jeong, kjyoon\}@kaist.ac.kr}\\
\url{https://vi.kaist.ac.kr/}}

\maketitle

\begin{abstract}
    Robust 3D object detection in adverse weather conditions is challenging due to sensor limitations. Although combining complementary modalities such as LiDAR and 4D RADAR has shown promise, the sparsity of these sensors becomes apparent in adverse weather with reduced reflections, leading to objects with few or no point cloud returns. To address this limitation, camera sensors provide visual cues even when LiDAR and RADAR signals are weakened. However, cameras themselves are also vulnerable to adverse weather, where some regions become unreliable due to snow or rain occluding the camera lens. While some camera-fusion methods designed for adverse weather learn to weigh image regions via confidence maps, these maps receive no direct supervision and are learned solely through the detection loss. We introduce Reliability-Aware Fusion (RAF), which explicitly supervises per-pixel reliability estimation and provides a direct learning signal for identifying and suppressing unreliable visual cues. Our framework leverages pretrained LiDAR--RADAR networks, keeping their backbones frozen while only training the added camera branch, BEV fusion encoder, and detection head. Extensive experiments on the K-Radar and VoD datasets demonstrate that integrating RAF consistently improves detection accuracy over LiDAR--RADAR baselines, achieving up to +6.5 $AP_{BEV}$ and +7.4 $AP_{3D}$ gains. Code is available at \href{https://github.com/parkie0517/RAF}{https://github.com/parkie0517/RAF}.
\end{abstract}

%%%%%%%%%%%%%%%%%%%%%%%%%%%%%%%%%%%%%%%%%%%%%%%%%%%%%%%%%%%%%%%%%%%%%%
%%%%%%%%%%%%%%%%%%%%%%%%%%% START OF INTRO %%%%%%%%%%%%%%%%%%%%%%%%%%%
%%%%%%%%%%%%%%%%%%%%%%%%%%%%%%%%%%%%%%%%%%%%%%%%%%%%%%%%%%%%%%%%%%%%%%

\section{Introduction}
\label{sec:intro}

\begin{figure}[tb]
  \centering
  \begin{subfigure}{0.49\linewidth}
    \includegraphics[width=\linewidth]{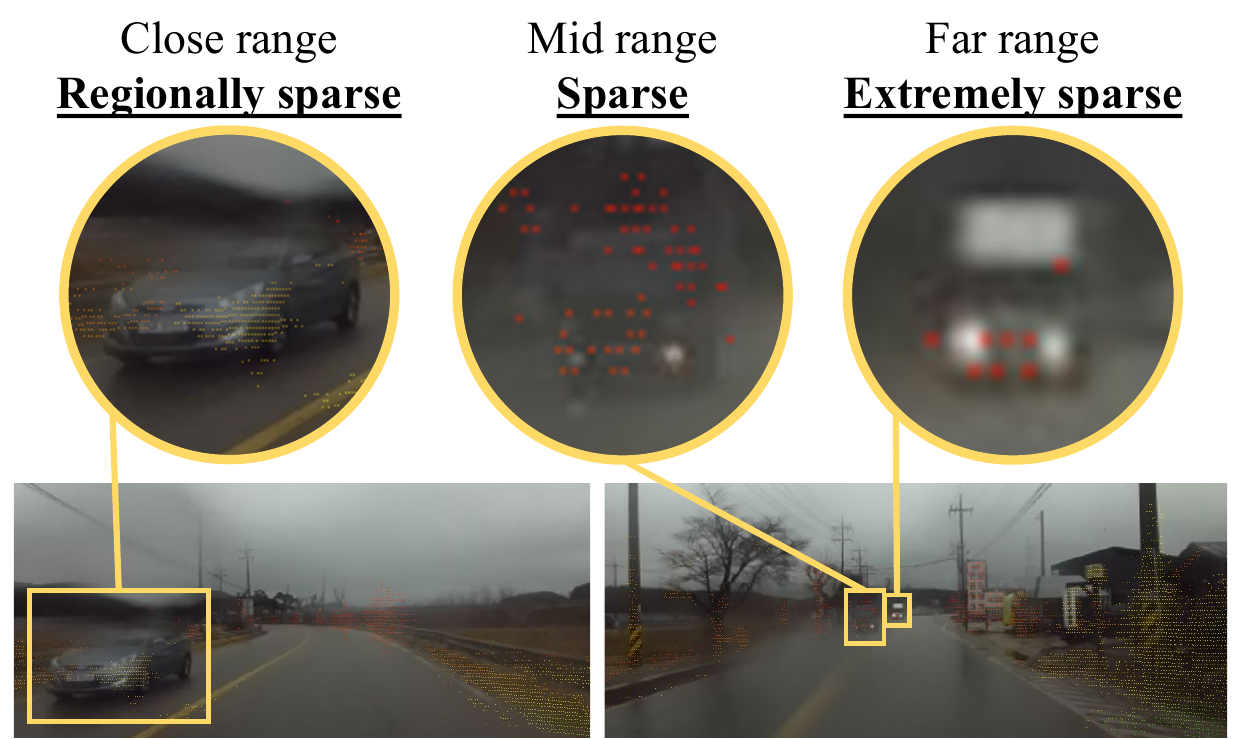}
    \caption{Sparse Point Cloud in Adverse Weather}
  \end{subfigure}
  \hfill
  \begin{subfigure}{0.49\linewidth}
    \includegraphics[width=\linewidth]{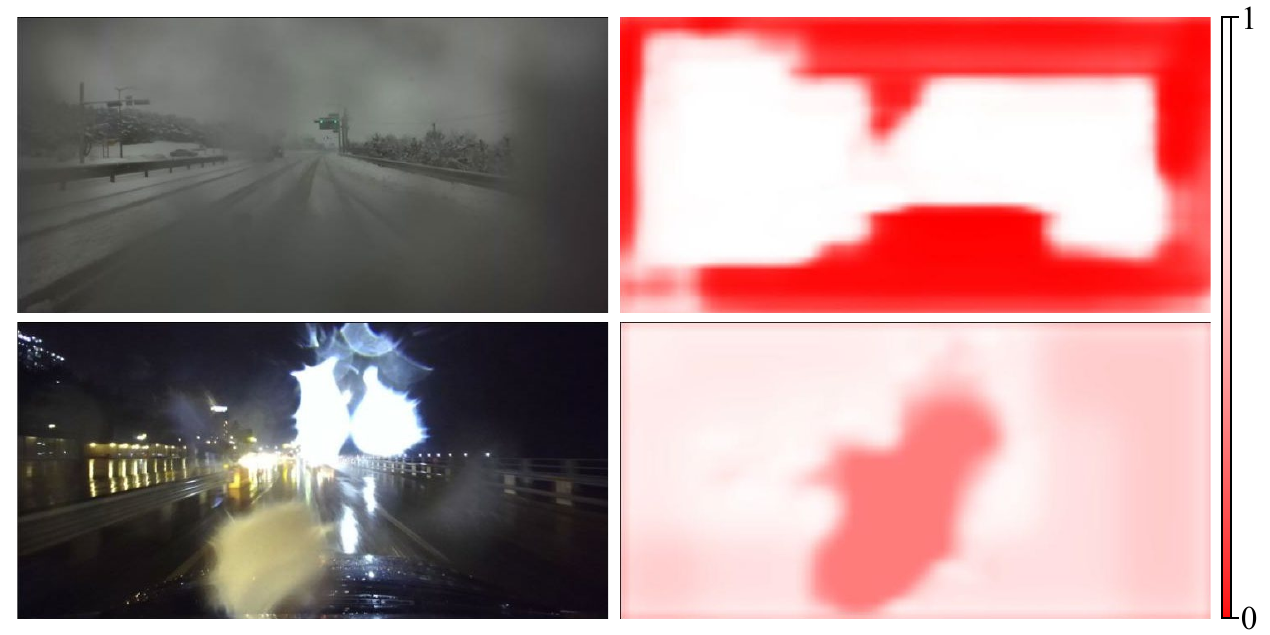}
    \caption{Camera Input Reliability Awareness}
  \end{subfigure}
  \caption{
  (a) LiDAR and RADAR points become highly sparse in adverse weather. Only LiDAR returns are projected, since RADAR points are difficult to visualize and are instead shown in the supplementary material. (b) Weather-induced noise appears in spatially uneven patterns across camera images. Our Reliability-Aware Fusion (RAF) estimates per-pixel reliability and effectively identifies unreliable image regions (red).
  }
  \label{fig:teaser}
\end{figure}

Perception in autonomous driving is crucial, as it allows vehicles to accurately understand the surrounding environment \cite{fayyad2020deep, huang2022multi, yeong2021sensor, xu2025survey} and provide useful information for planning \cite{teng2023motion, reda2024path}. Among various perception tasks, 3D object detection \cite{arnold2019survey, mao20233d, qian20223d} plays a central role by identifying and localizing objects in 3D space. Driven by large-scale benchmarks such as KITTI, Waymo Open Dataset, and nuScenes \cite{geiger2012we, geiger2013vision, sun2020scalability, caesar2020nuscenes, wilson2023argoverse}, as well as model architectures, 3D object detection has made remarkable progress.

Single-modality approaches have pioneered 3D object detection through specialized sensing capabilities. Camera-only detectors \cite{huang2021bevdet, harley2023simple, li2023bevdepth, wang2022detr3d, philion2020lift, Li2022BEVFormer, liu2022petr} provide dense and semantically rich information. However, they lack explicit depth information, making accurate 3D localization challenging. LiDAR-only methods \cite{zhou2018voxelnet, lang2019pointpillars, yin2021center, yan2018second, shi2020pv, li2021lidar, chen2023voxelnext} offer accurate geometric structure and distance estimation, forming the backbone of most high-performance 3D detectors. However, LiDAR density decreases with distance \cite{yang2024improving}, limiting its ability to detect small or distant objects. As a result, relying on a single sensor leaves detectors vulnerable to missing critical cues.

To overcome the inherent weaknesses of single sensors, multi-modal fusion has become a cornerstone of modern 3D object detection \cite{wang2023multi, wang2023multi_survey_and_tax, wang2023multi_sensor_fusion}. Early camera--LiDAR detectors achieved strong accuracy by combining LiDAR geometry with camera semantics \cite{bai2022transfusion, li2022deepfusion, liang2022bevfusion, liu2022bevfusion}. Camera--RADAR setups leveraged the complementary depth and Doppler cues of RADAR for motion-aware perception \cite{nabati2021centerfusion, lin2024rcbevdet, kim2023craft, zheng2023rcfusion}. Yet real-world autonomy demands robustness not only under ideal conditions but also under adverse weather, where rain, fog, and snow induce severe signal degradation. To study this, new datasets such as Seeing Through Fog \cite{bijelic2020seeing}, Place3D \cite{li2024your}, Robo3D \cite{kong2023robo3d}, 3D-C \cite{dong2023benchmarking}, and K-Radar \cite{paek2022k} emerged to provide data across diverse weather conditions.

Recent studies under adverse conditions have revisited single-modality detectors, such as LiDAR-only methods \cite{lin2022improved, park2025no}. However, single-modality detectors are vulnerable to missing or noisy returns, motivating a trend toward LiDAR--RADAR fusion to leverage RADAR's weather-robust properties \cite{chae2024towards, huang2025l4dr, chae2025doppler}. While LiDAR--RADAR fusion improves robustness over single-modality approaches, both sensors still suffer from reduced returns under severe weather, often leaving only a few points on object surfaces as visualized in Fig.~\ref{fig:teaser}(a). Here, the camera can still provide valuable visual cues to compensate for missing returns.

Although cameras provide rich semantics, their reliability also deteriorates under rain streaks, fog, or snow. As shown in Fig.~\ref{fig:teaser}(b), certain regions of the image contain discriminative features, while adverse weather corrupts and acts as noise in other regions. When such unreliable regions dominate, fusing camera features can harm performance rather than help. We observe this in our experiments, where adding camera input to representative LiDAR--4D RADAR fusion baselines~\cite{chae2024towards, huang2025l4dr} yields limited or even negative gains under adverse weather, despite improvements in normal weather conditions (Tab.~\ref{tab:visibility}). This highlights that naive camera-fusion is insufficient under adverse weather, motivating fusion strategies that account for the varying reliability of camera features.

Recent camera-fusion methods learn confidence maps to weigh image regions differently \cite{yue2026roburcdet}. However, these maps receive no direct supervision regarding their reliability and are optimized solely through the detection loss. We experimented with this approach and found that without proper supervision, confidence maps can harm detection performance (Tab.~\ref{tab:component_ablation}).

To address this, we introduce Reliability-Aware Fusion (RAF), which explicitly supervises a per-pixel reliability map to suppress noise from weather-corrupted image regions while preserving discriminative cues from visible regions. Since no reliability ground truth exists, we construct image-level cleanliness labels (\textit{Clean}, \textit{Mixed}, \textit{Noisy}) based on the degree of camera occlusion. These labels train cross-modal feature alignment between pretrained frozen LiDAR--RADAR and camera representations, producing pseudo reliability labels that supervise the reliability map. RAF adopts a modular design that adds a camera branch to existing pretrained LiDAR--RADAR detectors, training only the camera branch, BEV fusion encoder, and detection head while keeping the pretrained backbone frozen.

RAF performs cross-modal feature alignment by projecting LiDAR--RADAR voxel features onto the image feature plane to compute feature similarity. However, even small calibration errors can cause voxels to project to incorrect pixel locations. To handle this, we propose Calibration-Aware Local Matching (CALM), which searches a local window around each projected pixel and selects the highest-similarity match. This provides tolerance to projection misalignment without additional learnable parameters.

In summary, our RAF framework performs robust 3D object detection under adverse 
weather with the following contributions:
\vspace{-10pt}
\begin{itemize}
    \item We introduce RAF, a weakly-supervised modular camera-fusion framework 
    that explicitly suppresses weather-corrupted image regions by predicting and 
    applying a per-pixel reliability map, while preserving discriminative cues 
    from visible regions.
    \item We propose Calibration-Aware Local Matching (CALM), a parameter-free 
    mechanism that searches within local windows around projected points to achieve 
    robust cross-modal feature alignment, mitigating the impact of calibration errors.
    \item Extensive experiments on K-Radar and VoD demonstrate that RAF consistently 
    improves over LiDAR--RADAR baselines, achieving up to +6.5 $AP_{BEV}$ and 
    +7.4 $AP_{3D}$ gains.
\end{itemize}

%%%%%%%%%%%%%%%%%%%%%%%%%%%%%%%%%%%%%%%%%%%%%%%%%%%%%%%%%%%%%%%%%%%%%%
%%%%%%%%%%%%%%%%%%%%%%%%%%%% END OF INTRO %%%%%%%%%%%%%%%%%%%%%%%%%%%%
%%%%%%%%%%%%%%%%%%%%%%%%%%%%%%%%%%%%%%%%%%%%%%%%%%%%%%%%%%%%%%%%%%%%%%

%%%%%%%%%%%%%%%%%%%%%%%%%%%%%%%%%%%%%%%%%%%%%%%%%%%%%%%%%%%%%%%%%%%%%%
%%%%%%%%%%%%%%%%%%%%%%%%%% START OF RELWORK %%%%%%%%%%%%%%%%%%%%%%%%%%
%%%%%%%%%%%%%%%%%%%%%%%%%%%%%%%%%%%%%%%%%%%%%%%%%%%%%%%%%%%%%%%%%%%%%%
\section{Related Work}

\subsection{Multi-Modal 3D Object Detection}
\vspace{-3pt}
\textbf{Camera--LiDAR fusion} approaches employ attention-based frameworks for cross-modal interaction \cite{bai2022transfusion, li2022deepfusion, zhang2024sparselif}, BEV-based methods for spatially consistent fusion \cite{liu2022bevfusion, liang2022bevfusion, song2024graphbev, li2024gafusion}, or a combination of both \cite{yin2024fusion}. Other recent directions include diffusion-based modeling \cite{le2024diffuser}, Mixture of Experts for handling sensor failures \cite{park2025resilient}, and prompt tuning with foundation models \cite{li2025pf3det}. For \textbf{Camera--RADAR fusion}, early attempts like \cite{nabati2021centerfusion} extended 2D RADAR point clouds into 3D pillars for fusion. More recent BEV-based attention frameworks \cite{zheng2023rcfusion, kim2023craft, lin2024rcbevdet, chu2025racformer} further strengthen feature correspondence between camera and RADAR domains. Additionally, \cite{zhao2024crkd} distills knowledge from a camera--LiDAR teacher, transferring learned priors to a camera--RADAR fusion network. For \textbf{LiDAR--RADAR fusion}, \cite{wang2023bi} pioneered this direction by integrating 3D RADAR and LiDAR features within the BEV space, later followed by \cite{song2024lirafusion}, which refined the spatial alignment process.

\subsection{3D Object Detection in Adverse Weather}
\vspace{-3pt}
\textbf{LiDAR-only} approaches \cite{lin2022improved, hahner2021fog, hahner2022lidar} aim to enhance feature extraction from point clouds despite noise and sparsity.
\textbf{RADAR-only} methods \cite{paek2022k, kong2024rtnh_plus} have demonstrated strong weather resilience.
\textbf{LiDAR--RADAR fusion} frameworks further exploit complementary modalities. \cite{qian2021robust} introduced late-fusion to integrate modality-specific features, while \cite{li2022modality, li2023st} incorporated teacher--student consistency and data augmentation to enhance cross-modal robustness. \cite{wang2022interfusion} demonstrated that lightweight BEV fusion can yield gains in adverse scenarios. 3D-LRF~\cite{chae2024towards} dynamically modulated 4D RADAR information flow based on weather cues, and L4DR~\cite{huang2025l4dr} combined early raw-signal fusion with BEV-level refinement.
\textbf{Camera--LiDAR--RADAR fusion} methods have extended fusion to all three modalities to improve robustness. \cite{paek2025availability} focused on handling missing modalities, while SAMFusion~\cite{palladin2024samfusion} further incorporated additional modalities such as NIR gated imaging.
However, none of these methods addresses spatially varying camera occlusions due to adverse weather.

\subsection{Reliability-Aware Fusion}
\vspace{-3pt}
\textbf{Scalar Modality Weighting} approaches \cite{zhang2024sparselif, cho2024cocoon, kim2023adaptive, sural2024contextualfusion} produce scalar weights to balance camera and LiDAR features in BEV space. \cite{chen2025hyperdimensional} extends this to camera--LiDAR--RADAR fusion with patch-wise scalar values. These weights are coarse-grained and do not account for spatially varying pixel-level degradation from adverse weather.
\textbf{Reliability Map Estimation} methods such as RobuRCDet~\cite{yue2026roburcdet} estimate a camera confidence map directly from camera features, using it to complementarily suppress camera and amplify RADAR contributions under degraded conditions.
Both approaches learn their fusion weights solely through the detection loss, without direct signal indicating which spatial regions are degraded. In contrast, RAF explicitly supervises the reliability map, providing a direct signal to localize and suppress weather-corrupted regions.

%%%%%%%%%%%%%%%%%%%%%%%%%%%%%%%%%%%%%%%%%%%%%%%%%%%%%%%%%%%%%%%%%%%%%%
%%%%%%%%%%%%%%%%%%%%%%%%%%% END OF RELWORK %%%%%%%%%%%%%%%%%%%%%%%%%%%
%%%%%%%%%%%%%%%%%%%%%%%%%%%%%%%%%%%%%%%%%%%%%%%%%%%%%%%%%%%%%%%%%%%%%%

%%%%%%%%%%%%%%%%%%%%%%%%%%%%%%%%%%%%%%%%%%%%%%%%%%%%%%%%%%%%%%%%%%%%%%
%%%%%%%%%%%%%%%%%%%%%%%%%% START OF METHOD %%%%%%%%%%%%%%%%%%%%%%%%%%%
%%%%%%%%%%%%%%%%%%%%%%%%%%%%%%%%%%%%%%%%%%%%%%%%%%%%%%%%%%%%%%%%%%%%%%
\section{Proposed Method}
\label{sec:methods}

\begin{figure}[tb]
  \centering
  \includegraphics[width=1.0\linewidth]{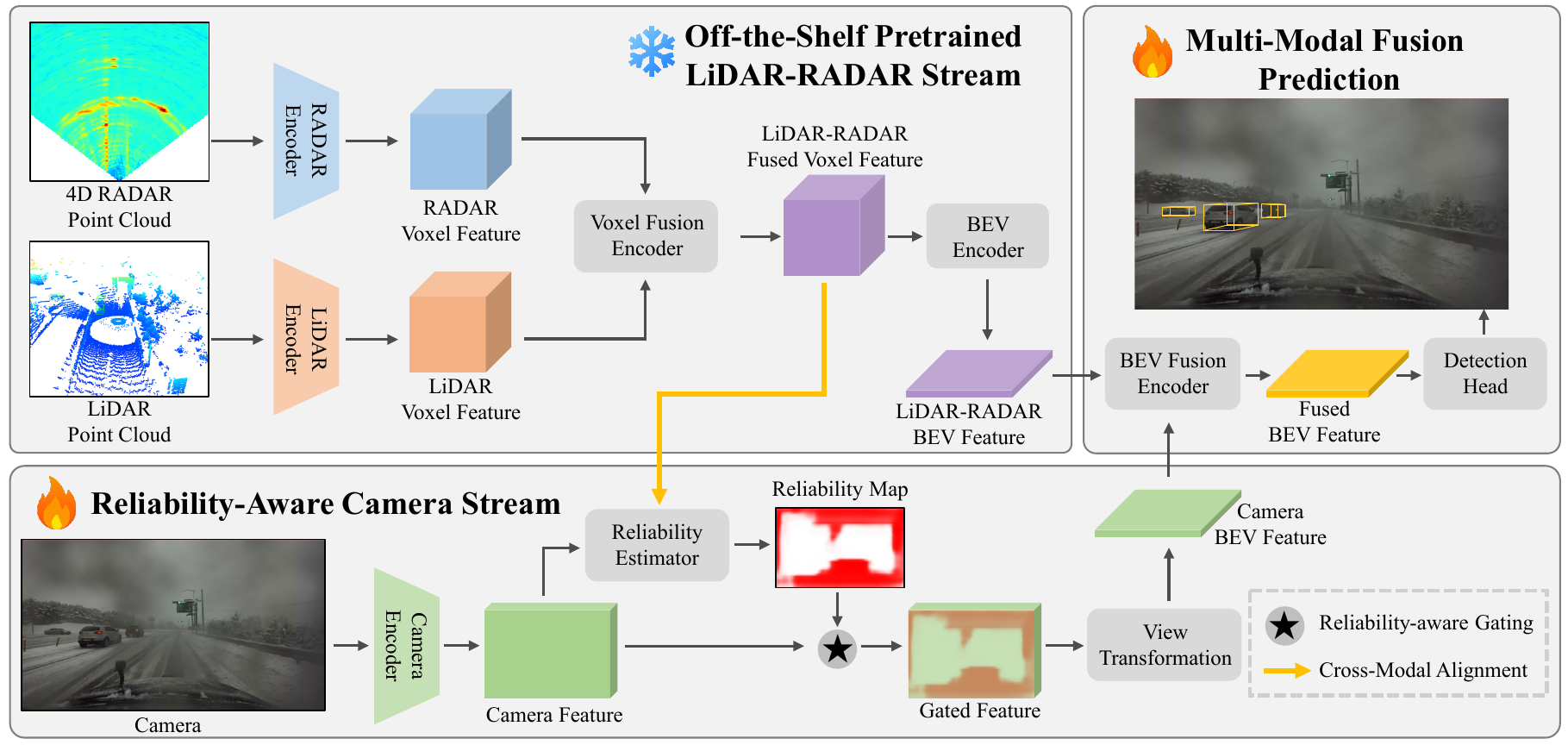}
  \caption{\textbf{Pipeline of our method.} Voxel features are extracted from a pretrained LiDAR--RADAR backbone, and camera features are modulated based on predicted reliability.}
  \label{fig:overview}
\end{figure}

\subsection{Overview}
\label{sec:overview}
As illustrated in Fig.~\ref{fig:overview}, RAF builds upon a pretrained LiDAR--RADAR detector and introduces a camera stream that predicts a per-pixel reliability map to gate image features. We freeze the LiDAR--RADAR backbone and train only the camera stream, BEV fusion encoder, and detection head. RAF consists of three key components: (1) a cross-modal similarity computation that produces pseudo reliability labels through image-level supervision, (2) Calibration-Aware Local Matching (CALM) that provides robustness to projection misalignment from calibration errors, and (3) sparse supervision that trains the reliability estimator from the pseudo reliability labels. At inference, only the reliability map prediction is needed without cross-modal similarity computation, CALM, or sparse supervision.

\begin{figure}[tb]
  \centering
  \includegraphics[height=7.0cm]{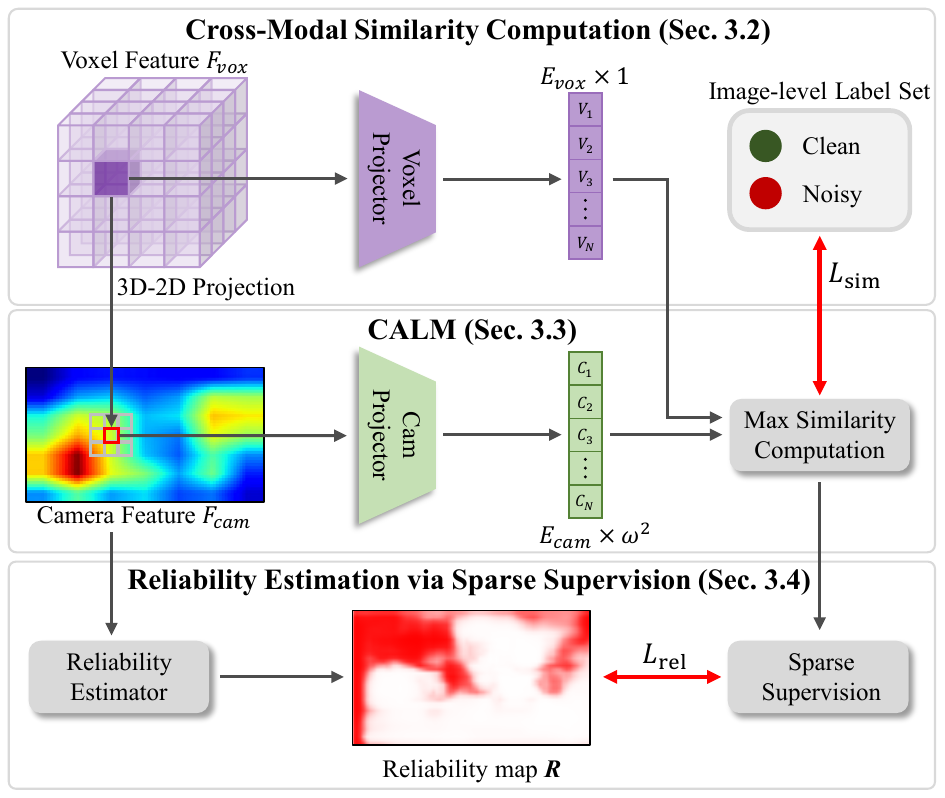}
  \caption{
      Overview of the training pipeline for reliability estimation. The top part (Sec.~\ref{sec:cross_modal_similarity_computation}) computes cross-modal similarity between voxel and camera features. The middle part (Sec.~\ref{sec:calm}) applies CALM to select the best match within a local window. The bottom part (Sec.~\ref{sec:reliability_estimation_via_sparse_supervision}) constructs sparse pseudo labels from the similarity scores and supervises the reliability estimator. The Cam Projector is introduced in Sec.~\ref{sec:cross_modal_similarity_computation} and reused in CALM. The CALM window is illustrated with $w = 3$ for visualization.
  }
  \label{fig:uem}
\end{figure}

\subsection{Cross-Modal Similarity Computation}
\label{sec:cross_modal_similarity_computation}

As shown in Fig.~\ref{fig:uem}, to estimate per-pixel reliability, RAF first computes a similarity score between each LiDAR--RADAR voxel feature and its corresponding image feature. This requires projecting valid voxels onto the image feature map using the calibration matrices provided by each dataset~\cite{paek2022k, palffy2022multi}.

\noindent\textbf{Latent Space Projection.}
The sampled camera and voxel features are mapped to a shared latent space through MLP-based projectors:
\begin{equation}
    \begin{aligned}
        \bm{E}_{\text{vox}}^{C_{\text{latent}} \times M} = \text{Proj}_{\text{vox}}(\mathcal{F}_{\text{vox}}^{C_{\text{vox}} \times M}),
        \\[0.5em]
        \bm{E}_{\text{cam}}^{C_{\text{latent}} \times M} = \text{Proj}_{\text{cam}}(F_{\text{cam}}^{C_{\text{cam}} \times M}),
    \end{aligned}
\end{equation}
producing embeddings $\bm{E}_{\text{vox}}, \bm{E}_{\text{cam}}$, both of dimension ${C_{\text{latent}} \times M}$, where $M$ is the number of valid voxel--pixel correspondences.

\noindent\textbf{Similarity Computation.}
The cosine similarity between each voxel--pixel pair is computed and normalized to $[0,1]$:
\begin{equation}
\bm{S} = \frac{1}{2} \left(\frac{\bm{E}_{\text{vox}} \cdot \bm{E}_{\text{cam}}}{||\bm{E}_{\text{vox}}||~||\bm{E}_{\text{cam}}||} + 1\right).
\end{equation}

\noindent\textbf{Image-Level Cleanliness Labels.}
Each sample is assigned one of three labels based on camera visibility. \textit{Clean} denotes an entirely unoccluded image. \textit{Noisy} denotes an image fully occluded by snow, rain, or fog. \textit{Mixed} denotes a partially occluded image (Fig.~\ref{fig:qual_reliability_map}).

\noindent\textbf{Similarity Loss.}
The similarity loss supervises the projectors using image-level cleanliness labels:
\begin{equation}
\mathcal{L}_{\text{sim}} =
\frac{1}{|\mathcal{B}|}
\sum_{n \in \mathcal{B}}
\left[
\frac{1}{M_n}
\sum_{i=1}^{M_n}
\mathrm{BCE}\!\left(
S_{n,i},\,
y_n
\right)
\right],
\label{revised_L_sim}
\end{equation}
where $S_{n,i}$ denotes the similarity of the $i$-th valid voxel--pixel correspondence in image $n$. $y_n \in \{0,1\}$ is the image-level cleanliness label (\textit{Noisy} $\rightarrow 0$, \textit{Clean} $\rightarrow 1$), and $\mathcal{B}$ is the set of images in the current batch that are either \textit{Clean} or \textit{Noisy}. \textit{Mixed} samples are excluded from $\mathcal{L}_{\text{sim}}$. The loss is averaged first within each image and then across images. This supervision updates only the projector weights ($\text{Proj}_{\text{vox}}, \text{Proj}_{\text{cam}}$) and camera encoder, while the LiDAR--RADAR backbone remains frozen. Since \textit{Mixed} images contain both visible and occluded regions, the projectors trained to distinguish \textit{Clean} from \textit{Noisy} generalize to \textit{Mixed} samples, producing meaningful similarity scores across regions of varying occlusion.

\subsection{Calibration-Aware Local Matching}
\label{sec:calm}

The similarity computation in Sec.~\ref{sec:cross_modal_similarity_computation} assumes that each voxel projects to its true corresponding pixel, but small calibration errors shift the projected location, degrading the similarity signal. To handle this, CALM defines a $w \times w$ neighborhood centered at each projected pixel and computes the similarity for all pixels within the window:
\begin{equation}
S_i^{\text{CALM}} = \max_{j \in \mathcal{W}_i} \frac{1}{2} \left(\frac{\bm{E}_{\text{vox},i} \cdot \bm{E}_{\text{cam},j}}{||\bm{E}_{\text{vox},i}||~||\bm{E}_{\text{cam},j}||} + 1\right),
\end{equation}
where $\mathcal{W}_i$ is the set of valid pixel locations within the $w \times w$ window around the projected location of voxel $i$. The maximum similarity within the window is used as the final similarity score for the voxel--pixel pair. CALM introduces no additional learnable parameters and operates on the same projectors from Sec.~\ref{sec:cross_modal_similarity_computation}. When $w = 1$, CALM reduces to standard single-pixel matching.

\subsection{Reliability Estimation via Sparse Supervision}
\label{sec:reliability_estimation_via_sparse_supervision}

The similarity scores from Sec.~\ref{sec:calm} serve as pseudo reliability labels at projected pixel locations. When multiple voxels project onto the same pixel, their similarity scores are averaged. Pixels without voxel projection receive no label. Since similarity scores are computed for all three image-level label cases, pseudo labels exist for \textit{Clean}, \textit{Noisy}, and \textit{Mixed} samples. The resulting pseudo label map is sparse as only projected locations are supervised.

\noindent\textbf{Reliability Estimator.}
The input to the reliability estimator is a camera feature map of shape $[C_{\text{cam}}, H_{\text{feat}}, W_{\text{feat}}]$. The estimator consists of a $3 \times 3$ convolution followed by batch normalization and ReLU, then a $1 \times 1$ convolution that projects to a scalar value. A sigmoid activation constrains the output to $[0, 1]$, producing the reliability map $\bm{R}$ of shape $[1, H_{\text{feat}}, W_{\text{feat}}]$, where higher values indicate more reliable visual cues. We adopt a CNN design over a transformer-based alternative because reliability estimation depends mainly on local spatial degradation patterns (\textit{e.g.}, blur, scattering, visibility loss), for which convolutions are effective and lightweight. A comparison of CNN and transformer-based reliability estimator design is provided in Sec.~\ref{sec:computational_efficiency}.

\noindent\textbf{Reliability Loss.}
The reliability loss $\mathcal{L}_{\text{rel}}$ is defined as a binary cross-entropy between the predicted reliability values $\bm{R}_{n,i}$ and the pseudo labels $\bm{S}_{n,i}$ at projected locations:
\begin{equation}
\mathcal{L}_{\text{rel}} =
\frac{1}{|\mathcal{B}|}
\sum_{n \in \mathcal{B}}
\left[
\frac{1}{M_n}
\sum_{i=1}^{M_n}
\mathrm{BCE}\!\left(
\bm{R}_{n,i},\, \bm{S}_{n,i}
\right)
\right],
\end{equation}
where both $\bm{R}_{n,i}$ and $\bm{S}_{n,i}$ lie within $[0,1]$. Although supervision is sparse, the convolutional structure of the reliability estimator produces a dense reliability map.

\noindent\textbf{Reliability Gating.}
The predicted reliability map gates the camera features before view transformation:
\begin{equation}
    \mathcal{F}_{\text{gated}} = \mathcal{F}_{\text{cam}} \odot (\alpha + (1 - \alpha) \bm{R}),
    \label{gating}
\end{equation}
where $\mathcal{F}_{\text{cam}}$ is the input camera feature map, $\bm{R}$ the predicted reliability map, $\odot$ the Hadamard product, and $\alpha \in [0, 1]$ a hyperparameter controlling retention strength. When $\alpha = 0$, unreliable pixels ($\bm{R}=0$) are fully suppressed and reliable ones ($\bm{R}=1$) are fully preserved. When $\alpha = 1$, no modulation is applied. The gated features $\mathcal{F}_{\text{gated}}$ are passed to the view transformation module. Ablation studies on $\alpha$ are provided in Sec.~\ref{sec:alpha_ablation}.

\subsection{Training and Inference}
\label{sec:training_and_inference}

\noindent\textbf{Training.}
The total training objective is:
\begin{equation}
\mathcal{L}_{\text{total}} = 
\mathcal{L}_{\text{det}} + 
\beta \, \mathcal{L}_{\text{sim}} + 
\gamma \, \mathcal{L}_{\text{rel}},
\end{equation}
where $\mathcal{L}_{\text{det}}$ is the detection loss, and $\beta$, $\gamma$ are weighting coefficients, which are ablated in Sec.~\ref{sec:loss_ablation}. Note that $\mathcal{L}_{\text{sim}}$ and $\mathcal{L}_{\text{rel}}$ are not independent. $\mathcal{L}_{\text{sim}}$ trains the projectors to produce meaningful similarity scores, which then serve as pseudo labels for $\mathcal{L}_{\text{rel}}$. Without $\mathcal{L}_{\text{sim}}$, the pseudo labels cannot be trained and $\mathcal{L}_{\text{rel}}$ receives no valid supervision. Conversely, removing $\mathcal{L}_{\text{rel}}$ means the similarity scores are never transferred to the reliability map. The LiDAR--RADAR backbone remains frozen, and only the camera backbone, projectors ($\text{Proj}_{\text{vox}}$, $\text{Proj}_{\text{cam}}$), reliability estimator, BEV fusion encoder, and detection head are trained.

\noindent\textbf{Inference.}
At inference, the reliability estimator directly produces the reliability map from camera features. No voxel-to-image projection, similarity computation, or CALM is required. The reliability map gates the camera features via Eq.~\ref{gating}, and the gated features are sent to the view transformation module.

%%%%%%%%%%%%%%%%%%%%%%%%%%%%%%%%%%%%%%%%%%%%%%%%%%%%%%%%%%%%%%%%%%%%%%
%%%%%%%%%%%%%%%%%%%%%%%%%% END OF METHOD %%%%%%%%%%%%%%%%%%%%%%%%%%%%%
%%%%%%%%%%%%%%%%%%%%%%%%%%%%%%%%%%%%%%%%%%%%%%%%%%%%%%%%%%%%%%%%%%%%%%

%%%%%%%%%%%%%%%%%%%%%%%%%%%%%%%%%%%%%%%%%%%%%%%%%%%%%%%%%%%%%%%%%%%%%%
%%%%%%%%%%%%%%%%%%%%%%%% START OF EXPERIMENTS %%%%%%%%%%%%%%%%%%%%%%%%
%%%%%%%%%%%%%%%%%%%%%%%%%%%%%%%%%%%%%%%%%%%%%%%%%%%%%%%%%%%%%%%%%%%%%%

\section{Experiments}

\subsection{Datasets and Evaluation}
\label{sec:datasets_and_evaluation}
\noindent\textbf{Datasets and Metrics.}
We conduct experiments on the K-Radar dataset~\cite{paek2022k}, which includes camera images with varying degrees of weather-induced occlusion. The dataset consists of 58 sequences with 17{,}458 training and 17{,}536 test samples collected under various weather conditions. The dataset includes synchronized camera, LiDAR, and 4D RADAR. Following the evaluation protocol of the original paper for 3D object detection, we report both $AP_{\text{BEV}}$ and $AP_{\text{3D}}$, evaluating on the Sedan class with IoU=$0.5$. We additionally evaluate on VoD~\cite{palffy2022multi}, with dataset details and results provided in the supplementary material.

\noindent\textbf{Image-Level Cleanliness Splits.}
K-Radar's original evaluation splits are based on weather type (\textit{e.g.}, Heavysnow, Fog, Rain), but do not capture the degree of camera degradation. We define three splits based on camera image quality: \textit{Clean}, \textit{Noisy}, and \textit{Mixed}. \textit{Clean} sequences have an entirely unoccluded camera image. LiDAR and RADAR are also clean since weather conditions are favorable, though non-weather noise such as 4D RADAR multipath reflections may still exist. \textit{Noisy} sequences have the entire image occluded by snow, rain, or fog, with no objects visible. LiDAR and RADAR visibility may also be degraded. \textit{Mixed} sequences contain images that have both occluded and visible regions. Of the 58 sequences in K-Radar, 21 are \textit{Clean}, 9 are \textit{Noisy}, and 28 are \textit{Mixed}.

\subsection{Training Setup}
\label{sec:training_setup}
\noindent\textbf{Implementation Details.}
We use a pretrained Swin-Tiny~\cite{liu2021swin} as the camera backbone with a GeneralizedLSSFPN~\cite{lin2017feature} neck. Input images are cropped by 80 pixels from the top and rescaled to $320 \times 640$. Camera features ($C_{\text{cam}} = 256$) are projected to BEV using Lift-Splat-Shoot (LSS) ~\cite{philion2020lift}. The shared embedding dimension is $C_{\text{latent}} = 128$. The reliability estimator follows the architecture described in Sec.~\ref{sec:reliability_estimation_via_sparse_supervision}. The point cloud range is set to $[0, 72]$, $[-6.4, 6.4]$, and $[-2, 6]$\,m along the X, Y, and Z axes. Default hyperparameters are $\alpha = 0.2$, $w = 5$, $\beta = 0.25$, and $\gamma = 0.25$. Ablations on $\alpha$, $w$, $\beta$, and $\gamma$ are provided in Sec.~\ref{sec:ablation}.

For 3D-LRF~\cite{chae2024towards}, the voxel size is $[0.4, 0.4, 0.4]$\,m. BEV fusion concatenates camera ($256$ channels) and LiDAR--RADAR ($1536$ channels) BEV features and applies a Conv2D-BN-ReLU module to produce a fused output of $1536$ channels. Training uses Adam with a learning rate of $0.001$, cosine annealing to $0.0001$, weight decay of $0.01$, and runs for 8 epochs. For L4DR~\cite{huang2025l4dr}, the voxel size is $[0.15, 0.16, 8.0]$\,m. The LSS output channel is $64$, which is encoded to $256$ via a Conv2D before BEV fusion. The BEV fusion structure is the same as 3D-LRF. Training uses AdamW with the same learning rate and scheduler, and runs for 20 epochs. For both models, the detection head input channel remains unchanged from the original configurations. All models are trained with batch size 4 on a single RTX 3090 (24GB).

\noindent\textbf{Training Variants.}
We define four variants to isolate the contribution of each design choice. \textit{Baseline} is the original LiDAR--RADAR network with no camera input~\cite{huang2025l4dr, chae2024towards}. \textit{Joint} adds a camera branch and trains all parameters from scratch. \textit{Frozen} uses a pretrained frozen LiDAR--RADAR backbone and trains only the camera branch, BEV fusion encoder, and detection head. It has the same total parameter count as Joint but fewer trainable parameters. \textit{RAF} follows the same frozen setup with the addition of MLP projectors and the reliability estimator, with CALM applied during training. Joint, Frozen, and RAF have similar total parameter counts, ensuring fair comparison. Baseline, Joint, and Frozen are trained with $\mathcal{L}_{\text{det}}$ only, while RAF is trained with $\mathcal{L}_{\text{det}} + \mathcal{L}_{\text{sim}} + \mathcal{L}_{\text{rel}}$. A detailed comparison of computational efficiency and detection performance across variants is provided in Sec.~\ref{sec:computational_efficiency}.

\begin{table}[t]
    \centering
    \caption{
        \textbf{3D object detection results on K-Radar.}
        C, L, and R denote camera, LiDAR, and 4D RADAR, respectively.
        $\dagger$ denotes methods reimplemented on K-Radar, as no official K-Radar results exist.
        $\ddagger$ denotes reproduced results that differ from the originally reported values.
    }
    \label{tab:main_results}
    \scriptsize
    \resizebox{1.0\textwidth}{!}{%
    \renewcommand{\arraystretch}{1.0}
    \setlength{\tabcolsep}{5pt}
    \begin{tabular}{c|c|c|ccccccc|c}
    \toprule
    Methods & Sensors & Metric & Normal & Overcast & Fog & Rain & Sleet & Lightsnow & Heavysnow & Total \\ 
    \hline\hline
    
    % SINGLE MODALITY
    % \multirow{2}{*}{RTNH \cite{paek2022k}} & \multirow{2}{*}{R} & $AP_{\text{BEV}}$ & 35.8 & 41.9 & 44.8 & 30.2 & 34.5 & 63.9 & 55.1 & 36.0 \\    
    %  &  & $AP_{\text{3D}}$ & 19.7 & 20.5 & 15.9 & 13.0 & 13.5 & 21.0 & 6.4 & 14.1 \\ 
    % \hline
    % \multirow{2}{*}{RTNH \cite{paek2022k}} & \multirow{2}{*}{L} & $AP_{\text{BEV}}$ & 65.4 & 87.4 & 83.8 & 73.7 & 48.8 & 78.5 & 48.1 & 66.3 \\ 
    %  &  & $AP_{\text{3D}}$ & 39.8 & 46.3 & 59.8 & 28.2 & 31.4 & 50.7 & 24.6 & 37.8 \\
    % \hline
    % \multirow{2}{*}{PointPillars \cite{lang2019pointpillars}} & \multirow{2}{*}{L} & $AP_{\text{BEV}}$ & 48.2 & 53.0 & 45.4 & 44.2 & 45.9 & 74.5 & 53.8 & 49.1 \\ 
    %  &  & $AP_{\text{3D}}$ & 21.8 & 28.0 & 28.2 & 27.2 & 22.6 & 23.2 & 12.9 & 22.4 \\ \hline
    
    % MULTI MODALITY
    % InterFusion (L+R)
    % \multirow{2}{*}{InterFusion \cite{wang2022interfusion}} & \multirow{2}{*}{L+R} & $AP_{\text{BEV}}$ & 50.0 & 59.0 & 80.3 & 50.0 & 22.7 & 72.2 & 53.3 & 52.9 \\ 
    %  &  & $AP_{\text{3D}}$ & 15.3 & 20.5 & 47.6 & 12.9 & 9.3 & 56.8 & 25.7 & 17.5 \\ \hline
    % CRN (C+R)
    \multirow{2}{*}{CRN$^\dagger$ \cite{kim2023crn}} & \multirow{2}{*}{C+R} & $AP_{\text{BEV}}$ & 39.9 & 54.6 & 66.3 & 40.0 & 32.4 & 57.4 & 40.4 & \cellcolor{gray!15}41.5 \\ 
     &  & $AP_{\text{3D}}$ & 25.0 & 31.7 & 44.8 & 25.5 & 20.3 & 31.9 & 28.6 & \cellcolor{gray!15}24.9 \\ \hline
    % Robu (C+R)
    \multirow{2}{*}{RobuRCDet$^\dagger$ \cite{yue2026roburcdet}} & \multirow{2}{*}{C+R} & $AP_{\text{BEV}}$ & 46.9 & 56.3 & 69.2 & 40.9 & 38.5 & 48.3 & 45.1 & \cellcolor{gray!15}46.7 \\ 
     &  & $AP_{\text{3D}}$ & 28.7 & 27.7 & 43.0 & 25.3 & 11.3 & 22.8 & 28.6 & \cellcolor{gray!15}27.2 \\
    \hline\hline
    % CRN (C+L+R)
    \multirow{2}{*}{CRN$^\dagger$ \cite{kim2023crn}} & \multirow{2}{*}{C+L+R} & $AP_{\text{BEV}}$ & 71.1 & 70.8 & 87.1 & 67.1 & 44.3 & 74.8 & 49.1 & \cellcolor{gray!15}66.4 \\ 
     &  & $AP_{\text{3D}}$ & 44.1 & 40.3 & 76.2 & 43.5 & 19.0 & 41.8 & 32.2 & \cellcolor{gray!15}41.4 \\ \hline
    % Robu (C+L+R)
    \multirow{2}{*}{RobuRCDet$^\dagger$ \cite{yue2026roburcdet}} & \multirow{2}{*}{C+L+R} & $AP_{\text{BEV}}$ & 62.2 & 76.3 & 88.6 & 58.3 & 55.0 & 76.8 & 60.0 & \cellcolor{gray!15}64.9 \\ 
     &  & $AP_{\text{3D}}$ & 40.4 & 48.8 & 63.6 & 41.7 & 41.3 & 48.4 & 38.9 & \cellcolor{gray!15}44.1 \\ \hline
    % SAMFusion (C+L+R)
    \multirow{2}{*}{SAMFusion$^\dagger$ \cite{palladin2024samfusion}} & \multirow{2}{*}{C+L+R} & $AP_{\text{BEV}}$ & 57.7 & 49.9 & 69.4 & 55.1 & 47.9 & 50.1 & 51.6 & \cellcolor{gray!15}56.7 \\ 
     &  & $AP_{\text{3D}}$ & 38.8 & 32.2 & 48.7 & 35.1 & 28.1 & 34.9 & 31.3 & \cellcolor{gray!15}36.8 \\
    \hline\hline

    % 3D-LRF COMPARISON
    % \multirow{2}{*}{3D-LRF \cite{chae2024towards}} & \multirow{2}{*}{L+R} & $AP_{\text{BEV}}$ & 72.3 & 88.4 & 86.6 & 76.6 & 47.5 & 79.6 & \textbf{64.1} & 73.6 \\ 
    %  &  & $AP_{\text{3D}}$ & 45.3 & 55.8 & 51.8 & 38.3 & 23.4 & \textbf{60.2} & 36.9 & 45.2 \\ \hline
    % 3D-LRF (baseline)
    \multirow{2}{*}{3D-LRF$^\ddagger$ (Baseline) \cite{chae2024towards}} & \multirow{2}{*}{L+R} & $AP_{\text{BEV}}$ & 62.1 & 88.5 & 81.8 & 69.9 & 37.7 & 84.6 & 36.0 & \cellcolor{gray!15}64.5 \\ 
     &  & $AP_{\text{3D}}$ & 32.9 & 49.1 & 40.4 & 34.5 & 25.1 & 53.1 & 14.8 & \cellcolor{gray!15}35.8 \\ \hline
    % 3D-LRF (joint)
    \multirow{2}{*}{3D-LRF (Joint)} & \multirow{2}{*}{C+L+R} & $AP_{\text{BEV}}$ & 70.2 & 81.9 & 75.8 & 74.5 & 18.4 & 74.2 & 40.9 & \cellcolor{gray!15}65.7 \\ 
     &  & $AP_{\text{3D}}$ & 41.3 & 61.3 & 49.0 & 39.2 & 25.5 & 59.0 & 27.7 & \cellcolor{gray!15}39.3 \\ \hline
    % 3D-LRF (frozen)
    \multirow{2}{*}{3D-LRF (Frozen)} & \multirow{2}{*}{C+L+R} & $AP_{\text{BEV}}$ & 67.2 & 87.8 & 81.6 & 78.3 & 42.4 & 80.1 & 48.6 & \cellcolor{gray!15}68.3 \\ 
     &  & $AP_{\text{3D}}$ & 38.2 & 57.2 & 44.7 & 36.5 & 25.0 & 55.6 & 39.5 & \cellcolor{gray!15}39.1 \\ \hline
    % 3D-LRF (RAF)
    \multirow{2}{*}{3D-LRF (RAF)} & \multirow{2}{*}{C+L+R} & $AP_{\text{BEV}}$ & \cellcolor{yellow!20}70.8 & \cellcolor{yellow!20}88.0 & \cellcolor{yellow!20}82.3 & \cellcolor{yellow!20}74.3 & \cellcolor{yellow!20}43.3 & \cellcolor{yellow!20}84.3 & \cellcolor{yellow!20}49.6 & \cellcolor{gray!15}71.0 \\ 
     &  & $AP_{\text{3D}}$ & \cellcolor{yellow!20}42.9 & \cellcolor{yellow!20}63.8 & \cellcolor{yellow!20}46.5 & \cellcolor{yellow!20}44.1 & \cellcolor{yellow!20}32.6 & \cellcolor{yellow!20}\textbf{63.9} & \cellcolor{yellow!20}35.6 & \cellcolor{gray!15}43.2 \\
    \hline\hline
     
    % L4DR COMPARISON
    % L4DR (baseline)
    \multirow{2}{*}{L4DR (Baseline) \cite{huang2025l4dr}} & \multirow{2}{*}{L+R} & $AP_{\text{BEV}}$ & 76.8 & \textbf{88.6} & 89.7 & 78.2 & 59.3 & 80.9 & 53.8 & \cellcolor{gray!15}77.5 \\ 
     &  & $AP_{\text{3D}}$ & 53.0 & 64.1 & 73.2 & 53.8 & 46.2 & 52.4 & 37.0 & \cellcolor{gray!15}53.5 \\ \hline
    % L4DR (joint)
    \multirow{2}{*}{L4DR (Joint)} & \multirow{2}{*}{C+L+R} & $AP_{\text{BEV}}$ & 77.0 & 80.8 & 89.4 & 78.4 & 59.9 & 81.1 & 53.2 & \cellcolor{gray!15}77.8  \\  
     &  & $AP_{\text{3D}}$ & 54.2 & 63.2 & 68.6 & 53.5 & \textbf{46.4} & 54.1 & 32.2 & \cellcolor{gray!15}53.9 \\ \hline
    % L4DR (frozen)
    \multirow{2}{*}{L4DR (Frozen)} & \multirow{2}{*}{C+L+R} & $AP_{\text{BEV}}$ & 79.8 & 83.5 & 92.2 & 79.1 & 59.2 & 85.7 & 58.4 & \cellcolor{gray!15}78.6 \\ 
     &  & $AP_{\text{3D}}$ & 58.2 & 53.6 & 71.5 & \textbf{55.9} & 36.6 & 46.6 & 35.3 & \cellcolor{gray!15}54.2 \\ \hline
    % L4DR (RAF)
    \multirow{2}{*}{L4DR (RAF)} & \multirow{2}{*}{C+L+R} & $AP_{\text{BEV}}$ & \cellcolor{yellow!20}\textbf{82.0} & \cellcolor{yellow!20}88.2 & \cellcolor{yellow!20}\textbf{93.9} & \cellcolor{yellow!20}\textbf{82.6} & \cellcolor{yellow!20}\textbf{63.0} & \cellcolor{yellow!20}\textbf{88.5} & \cellcolor{yellow!20}\textbf{60.4} & \cellcolor{gray!15}\textbf{82.0} \\ 
     &  & $AP_{\text{3D}}$ & \cellcolor{yellow!20}\textbf{59.4} & \cellcolor{yellow!20}\textbf{67.6} & \cellcolor{yellow!20}\textbf{79.3} & \cellcolor{yellow!20}53.3 & \cellcolor{yellow!20}45.9 & \cellcolor{yellow!20}56.3 & \cellcolor{yellow!20}\textbf{37.8} & \cellcolor{gray!15}\textbf{57.4} \\
    \bottomrule
    \end{tabular}
    }
    \vspace{-10pt}
\end{table}

\subsection{Main Results}
We compare against camera--RADAR (CRN~\cite{kim2023crn}, RobuRCDet~\cite{yue2026roburcdet}), camera--LiDAR--RADAR (SAMFusion~\cite{palladin2024samfusion}), and LiDAR--RADAR (3D-LRF~\cite{chae2024towards}, L4DR~\cite{huang2025l4dr}) fusion methods. To also evaluate CRN and RobuRCDet in the camera--LiDAR--RADAR setting, we add a sparse convolutional LiDAR encoder and concatenate LiDAR features before the BEV fusion encoder. As CRN, RobuRCDet, and SAMFusion assume 3D RADAR inputs without elevation, we discard the height dimension of the 4D RADAR point cloud to conform to their input format. Results are presented in Tab.~\ref{tab:main_results}.

Adding LiDAR to camera--RADAR methods yields large gains even under adverse weather (\textit{e.g.}, CRN $AP_{\text{BEV}}$: 41.5 $\rightarrow$ 66.4), confirming that LiDAR remains valuable in these conditions. Among all methods, L4DR (RAF) achieves the highest total score. Both L4DR (RAF) and 3D-LRF (RAF) achieve the best performance among their respective variants. Frozen generally performs on par or better than Joint. Under normal weather, adding camera consistently improves both L4DR and 3D-LRF over their baselines, which is expected since camera provides additional visual cues. However, under adverse weather conditions, the improvement is inconsistent. Adding camera sometimes improves and sometimes degrades performance. This indicates that weather-induced occlusions on camera images can act as noise.

Comparing Frozen and RAF against the Baseline reveals a clear difference. For L4DR (Frozen), 2 out of 7 weather conditions show lower $AP_{\text{BEV}}$ and 5 out of 7 show lower $AP_{\text{3D}}$ than the Baseline. In contrast, L4DR (RAF) shows lower $AP_{\text{BEV}}$ in only 1 out of 7 and lower $AP_{\text{3D}}$ in 2 out of 7. A similar pattern holds for 3D-LRF variants. By incorporating RAF, camera fusion becomes less susceptible to noise from weather-corrupted images.

\begin{table}[t]
    \centering
        \caption{
            \textbf{Performance comparison across different visibility.}
            Left: L4DR variants.
            Right: 3D-LRF variants.
            \textit{Clean}: no image degradation.
            \textit{Mixed}: partial occlusion from weather-induced noise.
            \textit{Noisy}: image entirely occluded with no informative features.
            $\ddagger$ denotes reproduced results.
        }
    \label{tab:visibility}
    
    \begin{minipage}{0.48\textwidth}
        \centering
        \resizebox{\textwidth}{!}{%
        \renewcommand{\arraystretch}{1.15}
        \setlength{\tabcolsep}{5pt}
        \begin{tabular}{c|c|c|ccc}
        \toprule
        Methods & Sensors & Metric & \textit{Clean} & \textit{Mixed} & \textit{Noisy} \\ 
        \hline\hline
        \multirow{2}{*}{L4DR (Baseline) \cite{huang2025l4dr}} & \multirow{2}{*}{L+R} 
         & $AP_{\text{BEV}}$ & 77.0 & 78.4 & \textbf{59.0} \\ 
         &  & $AP_{\text{3D}}$ & 50.9 & 54.9 & 43.1  \\
        \hline
        \multirow{2}{*}{L4DR (Joint)} & \multirow{2}{*}{C+L+R} 
         & $AP_{\text{BEV}}$ & \textbf{81.5} & 77.5 & 55.6 \\ 
         &  & $AP_{\text{3D}}$ & 55.3 & 53.8 & 38.3 \\ 
        \hline
        \multirow{2}{*}{L4DR (Frozen)} & \multirow{2}{*}{C+L+R} 
         & $AP_{\text{BEV}}$ & 80.2 & 78.1 & 55.6  \\ 
         &  & $AP_{\text{3D}}$ & 55.9 & 54.2 & 42.7  \\ 
        \hline
        \multirow{2}{*}{L4DR (RAF)} & \multirow{2}{*}{C+L+R} 
         & $AP_{\text{BEV}}$ & \cellcolor{yellow!20}80.4 & \cellcolor{yellow!20}\textbf{82.1} & \cellcolor{yellow!20}\textbf{59.0} \\ 
         &  & $AP_{\text{3D}}$ & \cellcolor{yellow!20}\textbf{58.0} & \cellcolor{yellow!20}\textbf{57.5} & \cellcolor{yellow!20}\textbf{43.6} \\ 
        \bottomrule
        \end{tabular}
        }
    \end{minipage}%
    \hfill
    \begin{minipage}{0.48\textwidth}
        \centering
        \resizebox{\textwidth}{!}{%
        \renewcommand{\arraystretch}{1.15}
        \setlength{\tabcolsep}{5pt}
        \begin{tabular}{c|c|c|ccc}
        \toprule
        Methods & Sensors & Metric & \textit{Clean} & \textit{Mixed} & \textit{Noisy} \\ 
        \hline\hline
        \multirow{2}{*}{3D-LRF$^\ddagger$ (Baseline) \cite{chae2024towards}} & \multirow{2}{*}{L+R} 
         & $AP_{\text{BEV}}$ & 68.8 & 73.0 &  \textbf{38.8} \\ 
         &  & $AP_{\text{3D}}$ & 41.6 & 43.5 & \textbf{29.1} \\ 
        \hline
        \multirow{2}{*}{3D-LRF (Joint)} & \multirow{2}{*}{C+L+R} 
         & $AP_{\text{BEV}}$ & 69.8 & 68.8 & 25.4 \\ 
         &  & $AP_{\text{3D}}$ & 43.8 & 40.3 & 24.1 \\ 
        \hline
        \multirow{2}{*}{3D-LRF (Frozen)} & \multirow{2}{*}{C+L+R} 
         & $AP_{\text{BEV}}$ & 70.4 & 65.5 & 30.8 \\ 
         &  & $AP_{\text{3D}}$ & 44.7 & 40.7 & 18.1 \\ 
        \hline
        \multirow{2}{*}{3D-LRF (RAF)} & \multirow{2}{*}{C+L+R} 
         & $AP_{\text{BEV}}$ & \cellcolor{yellow!20}\textbf{72.2} & \cellcolor{yellow!20}\textbf{74.5} & \cellcolor{yellow!20}38.4 \\ 
         &  & $AP_{\text{3D}}$ & \cellcolor{yellow!20}\textbf{45.2} & \cellcolor{yellow!20}\textbf{46.5} & \cellcolor{yellow!20}\textbf{29.1} \\ 
        \bottomrule
        \end{tabular}
        }
    \end{minipage}
\end{table}

\subsection{Visibility Analysis}
\label{sec:visibility}

Tab.~\ref{tab:visibility} reports results under the visibility splits defined in Sec.~\ref{sec:datasets_and_evaluation}. Both L4DR and 3D-LRF variants show similar trends across the three splits. Under \textit{Clean} conditions, all camera variants improve over the Baseline, confirming that camera features are beneficial when the image is unoccluded. RAF's noise suppression generally does not sacrifice clean performance.

Under \textit{Mixed} conditions, where images contain both occluded and visible regions, Joint and Frozen drop compared to the Baseline. The drops are small for L4DR but large for 3D-LRF. In contrast, RAF improves over the Baseline in both cases. This indicates that RAF has learned to suppress noisy features from occluded regions while preserving useful cues from visible regions. Under \textit{Noisy} conditions, where the entire image is occluded, Joint and Frozen drop compared to the Baseline, indicating that the noisy camera images degrade detection performance. RAF is the least affected among the camera variants, performing close to the Baseline. Overall, when noisy regions dominate the image, camera fusion can hurt detection. RAF's reliability gating makes camera fusion less susceptible to this noise. We additionally show that RAF applied to CRN, RobuRCDet, and SAMFusion exhibits similar trends, with details provided in the supplementary material.

\subsection{Ablation Study}
\label{sec:ablation}

\begin{table}[t]
\centering
    \caption{
        \textbf{Component Ablation.}
        Left: L4DR variants.
        Right: 3D-LRF variants.
        Gating denotes a reliability estimator trained solely with $\mathcal{L}_{\text{det}}$.
        WS denotes weakly-supervised reliability estimation with $\mathcal{L}_{\text{sim}} + \mathcal{L}_{\text{rel}}$.
    }
\label{tab:component_ablation}
\renewcommand{\arraystretch}{1.2}
\setlength{\tabcolsep}{3pt}

\begin{minipage}{0.48\textwidth}
    \centering
    {\scriptsize
    \resizebox{\textwidth}{!}{%
    \begin{tabular}{c|cccc|cc}
    \toprule
    \textbf{Variant} & \textbf{Pretrained} & \textbf{Gating} & \textbf{WS} & \textbf{CALM} & \textbf{$AP_{\text{BEV}}$} & \textbf{$AP_{\text{3D}}$} \\
    \hline \hline
    (a) &            &            &            &            & 77.8          &         53.9  \\
    (b) & \checkmark &            &            &            & 78.6          &         54.2  \\ % frozen이랑 성능 동일
    (c) & \checkmark & \checkmark &            &            & 76.4          &         53.2  \\ % det loss only
    (d) & \checkmark & \checkmark & \checkmark &            & 80.2          &         55.4  \\
    (e) & \checkmark & \checkmark & \checkmark & \checkmark & \textbf{82.0} & \textbf{57.4} \\
    \bottomrule
    \end{tabular}}}
\end{minipage}%
\hfill
\begin{minipage}{0.48\textwidth}
    \centering
    {\scriptsize
    \resizebox{\textwidth}{!}{%
    \begin{tabular}{c|cccc|cc}
    \toprule
    \textbf{Variant} & \textbf{Pretrained} & \textbf{Gating} & \textbf{WS} & \textbf{CALM} & \textbf{$AP_{\text{BEV}}$} & \textbf{$AP_{\text{3D}}$} \\
    \hline \hline
    (a) &            &            &            &            &         65.7  &         39.3  \\ % joint랑 동일
    (b) & \checkmark &            &            &            &         68.3  &         39.1  \\ % frozen이랑 성능 동일
    (c) & \checkmark & \checkmark &            &            &         68.0  &         33.1  \\ % det loss only
    (d) & \checkmark & \checkmark & \checkmark &            &         70.5  &         41.9  \\
    (e) & \checkmark & \checkmark & \checkmark & \checkmark & \textbf{71.0} & \textbf{43.2} \\
    \bottomrule
    \end{tabular}}}
\end{minipage}
\end{table}

\noindent\textbf{Component Ablation.}
\label{sec:component_ablation}
Tab.~\ref{tab:component_ablation} presents ablation results on L4DR (left) and 3D-LRF (right). Variant (a) trains all parameters from scratch (Joint), and (b) uses a pretrained frozen LiDAR--RADAR backbone (Frozen). Using a pretrained backbone generally improves performance. Variant (c) adds gating without supervision, optimized only through $\mathcal{L}_{\text{det}}$. This drops performance compared to (b). For L4DR, it falls even below (a), and for 3D-LRF, $AP_{\text{3D}}$ drops by 6.0. This confirms that gating without proper supervision harms detection.

Variant (d) adds weakly-supervised reliability estimation (WS), which includes both $\mathcal{L}_{\text{sim}}$ and $\mathcal{L}_{\text{rel}}$. These two losses are not independent. $\mathcal{L}_{\text{sim}}$ helps produce pseudo labels that $\mathcal{L}_{\text{rel}}$ uses. Therefore, they cannot be ablated separately. With WS, performance improves substantially, outperforming (a)--(c) on both backbones. Variant (e) adds Calibration-Aware Local Matching (CALM), achieving the best scores overall. With no additional learnable parameters, CALM provides up to $+1.8$ $AP_{\text{BEV}}$ / $+2.0$ $AP_{\text{3D}}$ for L4DR and $+0.5$ $AP_{\text{BEV}}$ / $+1.3$ $AP_{\text{3D}}$ for 3D-LRF.

\begin{table}[t]
    \renewcommand{\arraystretch}{1.2}
    \setlength{\tabcolsep}{3pt}

    \begin{minipage}{0.48\textwidth}
        \centering
        \caption{
            \textbf{Ablation on the reliability suppression strength $\alpha$.}
            $\alpha = 0.0$ applies hard gating and $\alpha = 1.0$ applies no gating.
        }
        \label{tab:alpha_ablation}
        {\scriptsize
        \resizebox{\textwidth}{!}{%
        \begin{tabular}{c|c|cccccc}
            \toprule
            \multirow{2}{*}{Methods} & \multirow{2}{*}{Metric} & \multicolumn{6}{c}{$\alpha$} \\
            & & 0.0 & 0.2 & 0.4 & 0.6 & 0.8 & 1.0 \\
            \hline \hline
            \multirow{2}{*}{3D-LRF (RAF)} & $AP_{\text{BEV}}$ & 70.6 & \textbf{71.0} & 70.3 & 69.9 & 69.3 & 68.3 \\
            & $AP_{\text{3D}}$ & \textbf{45.0} & 43.2 & 43.5 & 43.5 & 43.2 & 39.1 \\
            \hline \hline
            \multirow{2}{*}{L4DR (RAF)} & $AP_{\text{BEV}}$ & 81.4 & \textbf{82.0} & 81.3 & 80.3 & 80.3 & 78.6 \\
            & $AP_{\text{3D}}$ & 58.5 & 57.4 & \textbf{58.6} & 57.1 & 57.3 & 54.2 \\
            \bottomrule
        \end{tabular}}}
    \end{minipage}%
    \hfill
    \begin{minipage}{0.48\textwidth}
        \centering
        \caption{
            \textbf{Ablation on the CALM window size $w$.}
            $w = 1$ reduces to standard single-pixel matching.
        }
        \label{tab:window_ablation}
        {\scriptsize
        \resizebox{\textwidth}{!}{%
        \begin{tabular}{c|c|cccc}
            \toprule
            \multirow{2}{*}{Methods} & \multirow{2}{*}{Metric} & \multicolumn{4}{c}{$w$} \\
            & & 1 & 3 & 5 & 7 \\
            \hline \hline
            \multirow{2}{*}{3D-LRF (RAF)} & $AP_{\text{BEV}}$ & 70.5 & 70.3 & \textbf{71.0} & 68.4 \\
            & $AP_{\text{3D}}$ & 41.9 & 42.9 & \textbf{43.2} & 39.8 \\
            \hline \hline
            \multirow{2}{*}{L4DR (RAF)} & $AP_{\text{BEV}}$ & 80.2 & 81.1 & \textbf{82.0} & 79.4 \\
            & $AP_{\text{3D}}$ & 55.4 & 57.1 & \textbf{57.4} & 55.4 \\
            \bottomrule
        \end{tabular}}}
    \end{minipage}
\end{table}

\noindent\textbf{Ablation on the Reliability Suppression Strength ($\alpha$).}
\label{sec:alpha_ablation}
Tab.~\ref{tab:alpha_ablation} reports results for varying $\alpha$, where $\alpha = 0.0$ applies hard gating and $\alpha = 1.0$ applies no gating (equivalent to Frozen). Both backbones achieve the highest $AP_{\text{BEV}}$ at $\alpha = 0.2$, while the highest $AP_{\text{3D}}$ occurs at $\alpha = 0.0$ for 3D-LRF and $\alpha = 0.4$ for L4DR. Performance is relatively stable across $0.0$--$0.4$ but drops toward $1.0$. Hard gating ($\alpha = 0.0$) already performs well, indicating the reliability map is accurate enough to fully suppress unreliable pixels. We use $\alpha = 0.2$ as default, as it achieves the highest $AP_{\text{BEV}}$ for both backbones while maintaining competitive $AP_{\text{3D}}$.

\noindent\textbf{Ablation on the Window Size $w$.}
\label{sec:window_ablation}
Tab.~\ref{tab:window_ablation} reports results for varying the CALM window size $w$, where $w = 1$ reduces to standard single-pixel matching. Performance generally increases from $w = 1$ to $w = 5$, with $w = 5$ achieving the best scores for both backbones. At $w = 7$, performance drops to or below $w = 1$, as a window this large matches to pixels far from the true correspondence, introducing incorrect feature associations. We use $w = 5$ as default. Supplementary material includes experiments under synthetic calibration noise, to validate CALM's robustness to calibration error.

\begin{table}[t]
\centering
\caption{
    \textbf{Comparison of computational efficiency, reliability estimator design, and detection performance.}
     Transformer denotes RAF with a Vision Transformer-based reliability estimator.
    $\ddagger$ denotes reproduced results.
}
\label{tab:computational_efficiency}
% \resizebox{\textwidth}{!}{%
\resizebox{0.7\textwidth}{!}{%
\begin{tabular}{c|c|cccc|cc}
\toprule
\multirow{2}{*}{Methods} & \multirow{2}{*}{Sensors} 
& \multicolumn{4}{c|}{Computation} 
& \multicolumn{2}{c}{Metric} \\
& & Params (M) & GFLOPs & VRAM (GB) & FPS & $AP_{\text{BEV}}$ & $AP_{\text{3D}}$ \\
\hline\hline
3D-LRF$^\ddagger$ (Baseline) \cite{chae2024towards}& L+R & 34.83 & 55.38 & 0.43 & 9.75 & 64.5 & 35.8 \\
\hline
3D-LRF (Joint) & C+L+R & 91.57 & 394.93 & 1.14 & 6.95 & 65.7 & 39.3 \\
\hline
3D-LRF (Frozen) & C+L+R & 91.57 & 394.93 & 1.14 & 6.95 & 68.3 & 39.1 \\
\hline
3D-LRF (RAF) & C+L+R & 91.96 & 396.86 & 1.15 & 6.95 & 71.0 & 43.2 \\
\hline
3D-LRF (Transformer) & C+L+R & 92.12 & 407.51 & 1.15 & 6.90 & 70.8 & 40.2 \\
\hline\hline
L4DR (Baseline) \cite{huang2025l4dr}& L+R & 55.83 & 140.07 & 0.51 & 5.36 & 77.5 & 53.5 \\
\hline
L4DR (Joint) & C+L+R & 108.82 & 284.79 & 0.84 & 4.29 & 77.8 & 53.9 \\
\hline
L4DR (Frozen) & C+L+R & 108.82 & 284.79 & 0.84 & 4.29 & 78.6 & 54.2 \\
\hline
L4DR (RAF) & C+L+R & 126.42 & 286.81 & 0.94 & 4.19 & 82.0 & \textbf{57.4} \\
\hline
L4DR (Transformer) & C+L+R & 126.58 & 287.81 & 1.08 & 4.14 & \textbf{82.2} & 56.4 \\
\bottomrule
\end{tabular}}
\end{table}

\noindent\textbf{Computational Efficiency and Reliability Estimator Design.}
\label{sec:computational_efficiency}
Tab.~\ref{tab:computational_efficiency} compares parameter counts, GFLOPs, VRAM usage, FPS, and detection performance across five variants for both LiDAR--RADAR backbones. Adding a camera branch increases the number of parameters, GFLOPs, and VRAM while reducing FPS.
Joint and Frozen have identical computational costs, as they share the same architecture. RAF adds MLP projectors and a reliability estimator, resulting in a small increase in parameters and GFLOPs with negligible impact on FPS. We also compare the CNN-based reliability estimator (RAF) with a Vision Transformer-based~\cite{dosovitskiy2020image} variant (Transformer). The transformer variant adds slightly more parameters and GFLOPs. With fewer parameters, the CNN-based estimator achieves higher $AP_{\text{3D}}$ by $+1.0$ for L4DR and $+3.0$ for 3D-LRF, while maintaining comparable $AP_{\text{BEV}}$. We attribute this to the local nature of weather-induced noise on camera images, which convolutional operations are well-suited to capture.

\begin{table}[t]
\centering
\caption{
    \textbf{Sensitivity of loss weights $\beta$ and $\gamma$.}
    Left: L4DR (RAF). Right: 3D-LRF (RAF).
}
\label{tab:loss_ablation}

\begin{minipage}{0.49\textwidth}
    \centering
    {\scriptsize
    \begin{tabular}{cc|cc}
    \toprule
    \multicolumn{2}{c|}{Weight} & \multicolumn{2}{c}{Metric} \\
    $\beta$ & $\gamma$ & $AP_{\text{BEV}}$ & $AP_{\text{3D}}$ \\
    \hline\hline
    0.0  & 0.0  & 76.4 & 53.2 \\ % L_rel, L_sim 없이 gating만 한 거랑 동일
    \hline
    0.1  & 0.1  & 81.3 & 57.2 \\
    \hline
    0.25 & 0.25 & \textbf{82.0} & \textbf{57.4} \\
    \hline
    0.5  & 0.5  & 81.2 & 56.8 \\
    \bottomrule
    \end{tabular}}
\end{minipage}%
\hfill
\begin{minipage}{0.49\textwidth}
    \centering
    {\scriptsize
    \begin{tabular}{cc|cc}
    \toprule
    \multicolumn{2}{c|}{Weight} & \multicolumn{2}{c}{Metric} \\
    $\beta$ & $\gamma$ & $AP_{\text{BEV}}$ & $AP_{\text{3D}}$ \\
    \hline\hline
    0.0  & 0.0  & 68.0 & 33.1 \\ % L_rel, L_sim 없이 gating만 한 거랑 동일
    \hline
    0.1  & 0.1  & \textbf{71.0} & 41.9 \\
    \hline
    0.25 & 0.25 & \textbf{71.0} & 43.2 \\
    \hline
    0.5  & 0.5  & 70.3 & \textbf{44.3} \\
    \bottomrule
    \end{tabular}}
\end{minipage}
\vspace{-5pt}
\end{table}

\noindent\textbf{Sensitivity of Loss Weights $\beta$ and $\gamma$.}
\label{sec:loss_ablation}
Tab.~\ref{tab:loss_ablation} reports results for varying $\beta$ and $\gamma$. Left: L4DR (RAF). Right: 3D-LRF (RAF). When both weights are $0.0$, only $\mathcal{L}_{\text{det}}$ is applied, reducing to gating without $\mathcal{L}_{\text{sim}}$ and $\mathcal{L}_{\text{rel}}$. The jump from $0.0$ to $0.1$ is large for both backbones, particularly for 3D-LRF $AP_{\text{3D}}$ ($33.1 \rightarrow 41.9$), but performance is relatively stable across $0.1$--$0.5$. This shows that having supervision matters significantly, while the exact weight is not critical. We use $\beta = \gamma = 0.25$ as default, as it achieves the best performance on L4DR and a balanced result on 3D-LRF.

\subsection{Qualitative Results}
\label{sec:qualitative_results}

\begin{figure}[tb]
  \centering
  \includegraphics[width=1.0\linewidth]{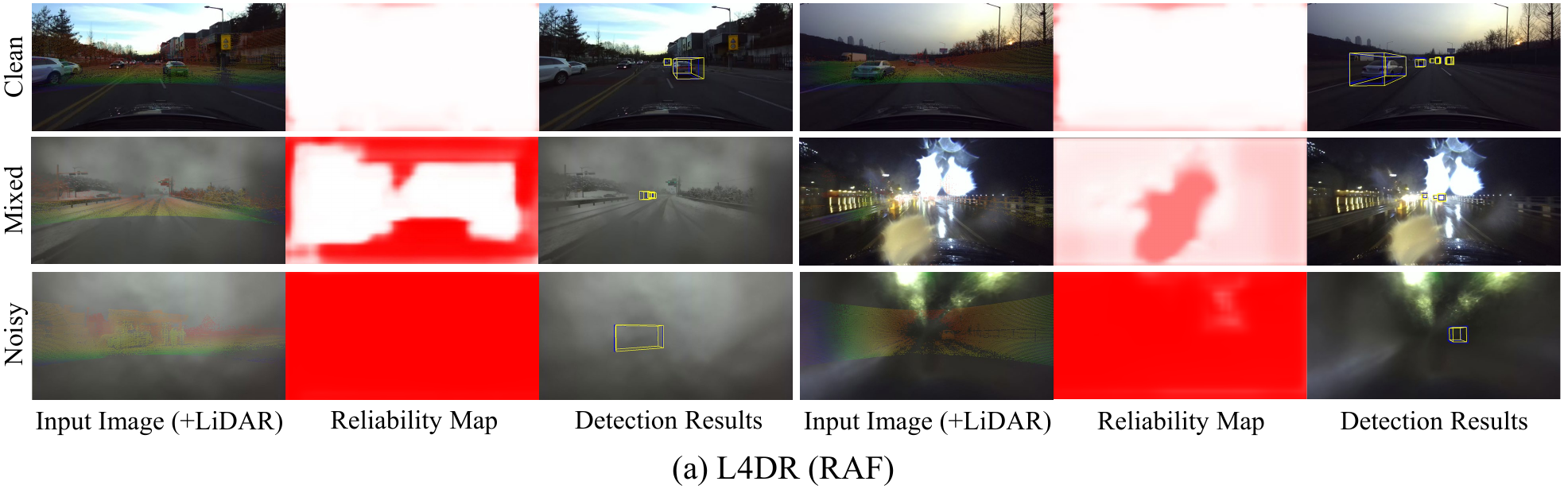} \\
  \includegraphics[width=1.0\linewidth]{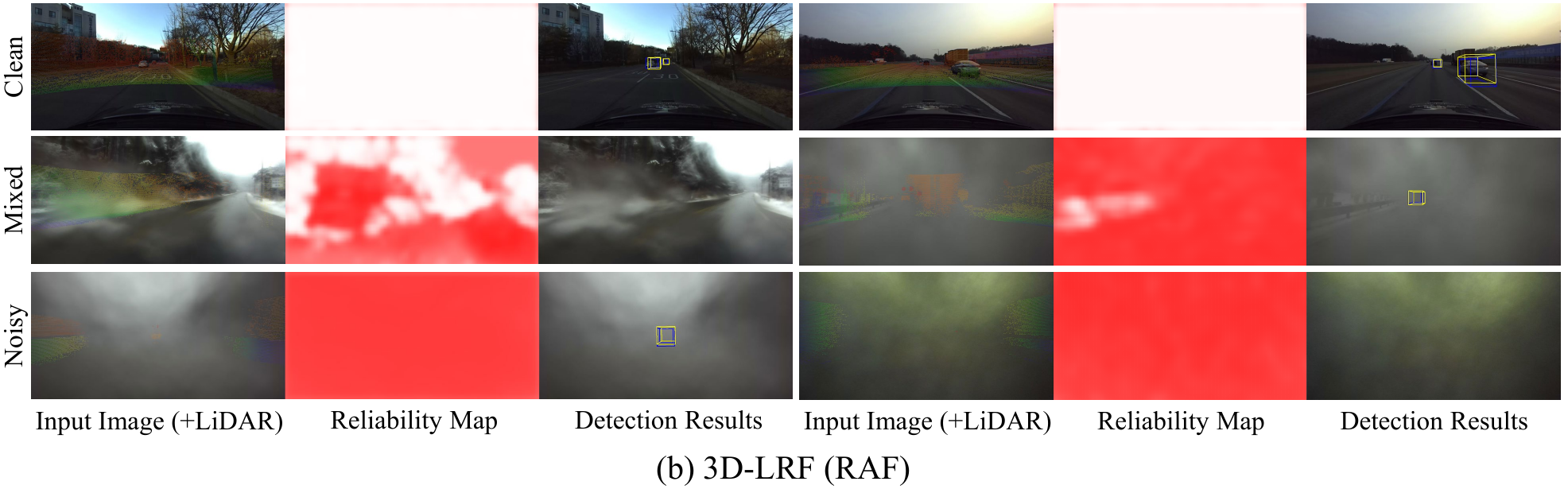}
  \caption{
        \textbf{Qualitative results under varying camera visibility.}
        \textit{Clean}: occlusion-free images.
        \textit{Mixed}: images containing both occluded and visible regions.
        \textit{Noisy}: images entirely occluded. 
        In the reliability maps, red indicates unreliable regions and white indicates reliable regions.
        Ground-truths are shown in blue and predictions in yellow.
  }
  \label{fig:qual_reliability_map}
  \vspace{-10pt}
\end{figure}

\vspace{-2.5pt}
\noindent\textbf{Estimated Camera Reliability.}
Fig.~\ref{fig:qual_reliability_map} shows that under \textit{Clean}, the reliability map remains uniformly white, indicating noise-free camera features. Under \textit{Mixed}, the model localizes corrupted regions (red) while preserving high reliability elsewhere. Under \textit{Noisy}, low reliability is correctly assigned across the entire image. These results demonstrate that our weak supervision accurately identifies weather-induced noise regions. Additional qualitative examples are provided in the supplementary material.

%%%%%%%%%%%%%%%%%%%%%%%%%%%%%%%%%%%%%%%%%%%%%%%%%%%%%%%%%%%%%%%%%%%%%%
%%%%%%%%%%%%%%%%%%%%%% END OF EXPERIMENTS %%%%%%%%%%%%%%%%%%%%%%%%%%%%
%%%%%%%%%%%%%%%%%%%%%%%%%%%%%%%%%%%%%%%%%%%%%%%%%%%%%%%%%%%%%%%%%%%%%%

\section{Conclusion}
\vspace{-2.5pt}
\label{sec:conclusion}
We proposed Reliability-Aware Fusion (RAF), a weakly-supervised framework that estimates per-pixel reliability to gate camera features. We also introduced Calibration-Aware Local Matching (CALM), which provides robustness to calibration errors without additional learnable parameters. RAF learns to suppress weather-corrupted camera features while preserving useful cues. Experiments on K-Radar demonstrate consistent overall gains over LiDAR--RADAR baselines and other camera-fusion approaches. RAF achieves these gains on two different LiDAR--RADAR backbones (L4DR and 3D-LRF), demonstrating its adaptability to other LiDAR--RADAR detectors. Future work includes extending reliability estimation to LiDAR and RADAR modalities, to address sensor degradation beyond cameras.

\section*{Acknowledgements}
This work was supported by the Technology Innovation Program (2410018126, KT224355, Development of autonomous driving connectivity technology based on sensor-infrastructure cooperation) funded By the Ministry of Trade, Industry \& Energy(MOTIE, Korea)
% \clearpage  % TODO FINAL: This \clearpage needs to be removed from both review and camera-ready versions.

\bibliographystyle{splncs04}
\bibliography{main}
\end{document}